%% file: main.tex
\documentclass{article}

\usepackage{amsmath,amsthm,amsfonts,amssymb,amscd}
\usepackage[a4paper,margin=25mm]{geometry}
\usepackage{booktabs,longtable,pdflscape,array,tabularx,microtype}
\usepackage{graphicx} 

\usepackage{gensymb}

\usepackage[dvipsnames]{xcolor}

\usepackage{listings}

\definecolor{codegreen}{rgb}{0,0.6,0}
\definecolor{codegray}{rgb}{0.5,0.5,0.5}
\definecolor{codepurple}{rgb}{0.58,0,0.82}
\definecolor{backcolour}{rgb}{0.95,0.95,0.92}
\lstdefinestyle{mystyle}{
    backgroundcolor=\color{backcolour},   
    commentstyle=\color{codegreen},
    keywordstyle=\color{magenta},
    numberstyle=\tiny\color{codegray},
    stringstyle=\color{codepurple},
    basicstyle=\ttfamily\footnotesize,
    breakatwhitespace=false,         
    breaklines=true,                 
    captionpos=b,                    
    keepspaces=true,                 
    numbers=left,                    
    numbersep=5pt,                  
    showspaces=false,                
    showstringspaces=false,
    showtabs=false,                  
    tabsize=2
}
\usepackage{textcomp}
\usepackage[T1]{fontenc}
\usepackage{lmodern}

\usepackage{indentfirst}

\usepackage{float}
\usepackage[section]{placeins}

\usepackage{mathtools}

\usepackage{general_commands}
\usepackage{marl_commands}
\usepackage{comment_commands}
\usepackage[hidelinks]{hyperref}

\redactionfalse

\showcommentsfalse

\title{Zero Collapse: A Failure Mode of Policy Gradient Methods in Discontinuous Reward Environments}
\author{ \redact{Nishant Kumar} \redact{Enrique Areyan Viqueira} \redact{Amy Greenwald} }
\date{} 

\begin{document}

\maketitle

\begin{abstract}
Policy-gradient learning can deteriorate even after substantial initial improvement. We study this behavior in a repeated first-price auction setting with a sharp threshold between winning and losing bids. We identify a failure mode, zero collapse, in which a policy shifts into persistently near-zero reward as informative outcomes become rarely sampled. Experiments with REINFORCE and a TD state-value actor-critic illustrate this failure and evaluate practical mitigations. Adaptive update-magnitude control and smooth-output parameterization support sustained REINFORCE learning, while a learned baseline substantially stabilizes the fixed-rate comparison. Some actor-critic runs still collapse under these interventions. A diagnostic near one observed collapse reveals severe continuation-value errors that reverse the relative desirability of actions across the winning threshold. These results illustrate how thresholded outcomes, policy updates, and inaccurate value estimates can contribute to zero collapse, and demonstrate practical mitigations in the studied setting.
\end{abstract}

\section{Introduction}
Bidding in repeated auctions provides a natural domain for reinforcement learning, combining continuous actions, sequential decisions, and strategic interaction among agents. Applications such as digital advertising \cite{zhang2014optimal} require bidders to balance the value of allocations against their cost. Policy-gradient methods \cite{williams1992reinforce,sutton2000policy} offer a direct approach to learning bidding policies, but the threshold governing whether a bid wins can create difficult learning dynamics.

In a first-price auction, bids below a competitor's bid lose, while bids above it win and incur payment. Crossing this threshold changes the allocation abruptly; further increases in a winning bid increase its cost. In the repeated-auction setting studied here, expenditure is incurred at each step and campaign value is awarded at termination. A policy that consistently bids below the winning threshold obtains neither impressions nor campaign value and receives zero cumulative reward. Policies that sometimes win encounter informative returns reflecting the value of the impressions obtained and the expenditure required.

We study a failure mode we term \emph{zero collapse}: after an initial period of learning, a policy shifts into behavior that yields persistently near-zero reward and little informative learning signal. Near the winning threshold, policy updates can move probability mass from profitable bids into the losing region. Once informative outcomes become rarely sampled, subsequent updates provide little signal for recovery. A policy may improve substantially before collapsing into the zero-reward region.

Our experiments illustrate zero collapse in REINFORCE and a TD state-value actor-critic. We examine initialization, adaptive update-magnitude control, smooth-output parameterization, and an action-independent baseline as practical mitigations. The interventions support sustained REINFORCE learning across multiple tested configurations, while some actor-critic runs still collapse after substantial improvement. A diagnostic near one actor-critic collapse reveals severe continuation-value errors that reverse the relative desirability of actions across the winning threshold. This illustrates how the actor can receive a misleading learning signal when it acts on valuations before they are sufficiently accurate.

Our contributions are the following:
\begin{itemize}
\item We identify and illustrate zero collapse in a repeated first-price auction setting.
\item We explain how movement into a zero-reward region can deprive learning of informative samples.
\item We diagnose continuation-value errors that reverse action desirability across the winning threshold near an observed actor-critic collapse.
\item We evaluate practical mitigations through fixed-rate controls, isolated interventions, and a combined treatment, with supporting parameterization and optimizer comparisons.
\end{itemize}
The appendix provides the auction formulation, policy-gradient derivations, experimental protocol, and supporting results.

\section{Related Work}

Policy gradient methods have long been a fundamental approach for reinforcement learning in continuous and stochastic environments. Early work such as REINFORCE demonstrated how gradient ascent on expected return can be performed directly through stochastic policies \cite{williams1992reinforce}. Subsequent work extended these ideas to actor-critic methods, which combine policy optimization with learned value-function approximations to reduce variance and improve sample efficiency \cite{sutton2000policy}.

Policy-update control is a related approach to stabilizing learning. Trust Region Policy Optimization (TRPO) \cite{schulman2015trpo} uses a trust-region constraint, while Proximal Policy Optimization (PPO) \cite{schulman2017ppo} uses a surrogate objective designed to discourage large policy changes. These methods motivate the future-work direction discussed below.

Our work is also related to the broader literature on sparse-reward and exploration challenges in reinforcement learning. In environments where informative reward is rare, policies may struggle to discover useful behaviors, motivating techniques such as intrinsic motivation, count-based exploration, and reward shaping \cite{pathak2017curiosity}. However, the phenomenon studied here differs from classical sparse-reward settings. In our case, informative reward may be readily discoverable, yet policies can still collapse due to the interaction between discontinuous reward geometry and policy optimization dynamics.

Finally, reinforcement learning has been applied extensively to auction and real-time bidding environments, particularly in digital advertising \cite{zhang2014optimal}. These domains naturally exhibit thresholded and discontinuous reward structures, making them a useful setting for studying the optimization challenges considered in this work.

Our focus is the empirical characterization of zero collapse in the studied auction setting, its observed learning dynamics, and practical mitigations.

\section{Reward Geometry of Thresholded Environments}
\label{sec:reward_geometry}
We begin with a simplified one-dimensional payoff function:
\[
r(a)=\begin{cases}0,&a<h,\\f(a),&a\geq h,\end{cases}
\]
where $a$ denotes the action, $h$ is a threshold, and $f(a)$ is decreasing with $f(h)>0$. The function consists of a flat zero-payoff region below the threshold and a non-flat region above it.

A one-step first-price auction provides a concrete example. Against a fixed competing bid $h$, a losing bid yields zero payoff, while a winning bid yields $v-a$, where $v>h$ is the value of the allocation. Thus, $f(a)=v-a$: the payoff jumps at the winning threshold and then decreases as the bid increases. For this schematic, we assign the threshold itself to the winning branch; the experimental environment resolves tied bids by random allocation.

\begin{figure}[!htbp]
\centering
\includegraphics[width=.78\linewidth]{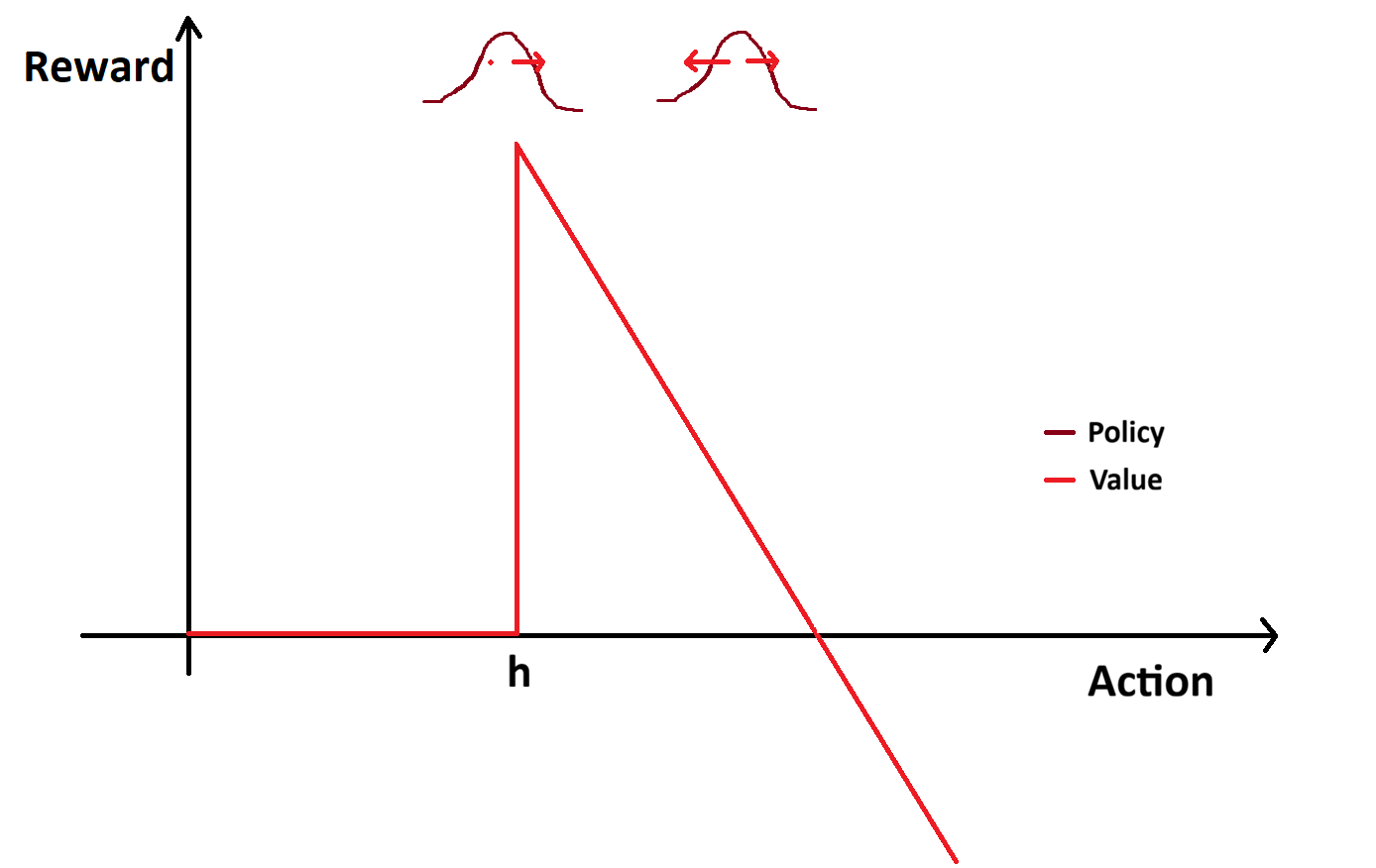}
\caption{Simplified payoff geometry for a one-step thresholded auction: losing bids yield zero payoff, while winning bids yield a payoff that decreases with the bid. The overlaid policy distribution illustrates how an update can move probability mass across the threshold into the zero-payoff region. The diagram illustrates the mechanism rather than the full action-value function of the repeated-auction experiment.}
\label{fig:reward_function_with_policy}
\end{figure}

As illustrated in Figure~\ref{fig:reward_function_with_policy}, this function has three key properties: a flat region where actions yield identical zero payoff, a discontinuous transition to positive payoff, and a decreasing branch whose highest payoff lies near the threshold. Together, these produce a cliff-like landscape in which a small change in action can move between profitable and zero-payoff outcomes.

The diagram isolates the geometry used to explain the learning mechanism. In the repeated-auction experiments, expenditure is incurred at each step and campaign value is awarded at termination. The schematic therefore represents a simplified net payoff rather than the immediate reward at every timestep. The same winning threshold governs allocation, and a policy that loses every auction receives zero cumulative reward. In the repeated game, losing one auction does not imply a zero return, because the policy may earn rewards in later auctions.

For a stochastic policy, the learning signal depends on how its action distribution overlaps these regions. When the policy samples the non-flat branch, observed payoffs can guide updates toward more profitable actions near the threshold. If an update instead moves most probability mass below the threshold, informative outcomes become rare. The next section examines how this loss of sampled signal can produce zero collapse.

\section{Zero Collapse: Mechanism and Dynamics}
\label{sec:mechanism}
We now examine how policy-gradient learning interacts with the geometry described in Section~\ref{sec:reward_geometry}. For REINFORCE, a sampled trajectory contributes an update direction of the form
\[
\widehat g(\theta)=\sum_{t=0}^{T-1}\nabla_\theta\log\pi_\theta(a_t\mid s_t)G_t,
\]
where $G_t$ is the sampled return-to-go. Each term favors increasing the likelihood of its sampled action for a positive return and decreasing it for a negative return. Learning therefore depends on the outcomes sampled by the current policy.

In the simplified one-step landscape, bids on the winning branch produce payoffs that vary with the bid. These samples provide information about profitable actions near the threshold. However, finite updates based on stochastic gradient estimates can shift the policy past this region, placing much of its probability mass below the winning threshold.

As winning actions become less likely, batches increasingly contain zero-return samples. For baseline-free REINFORCE, these samples contribute exactly zero to the update. If an entire batch has zero returns, its policy-gradient estimate is zero. When informative samples still occur but are rare, they provide only intermittent opportunities for the policy to move back toward profitable behavior.

This creates an asymmetry in the learning dynamics: an update supported by informative samples can move the policy into the losing region, but recovery requires sampling informative outcomes again. A policy may therefore improve substantially before collapsing into persistently near-zero reward. We call this failure mode \emph{zero collapse}, emphasizing the accompanying loss of sampled learning signal.

In the repeated-auction experiment, the same argument applies at the trajectory level. Losing a single auction need not eliminate the return from later auctions. When the policy instead loses throughout an episode, it incurs no expenditure and earns no campaign value, so all returns along that trajectory are zero. As such trajectories dominate the batch, the direct reward signal available to REINFORCE diminishes.

A learned baseline or critic changes the update signals in this regime. Their estimates can produce nonzero updates even when sampled returns are zero, but those updates need not guide the policy back toward profitable actions. In particular, baseline-adjusted contributions can fluctuate or cancel without supplying a systematic recovery direction. Section~\ref{sec:mitigation} examines the baseline's stabilizing role, and Section~\ref{sec:critic_mechanism} diagnoses misleading action valuations in the TD actor-critic.

Zero collapse does not require informative actions to become mathematically impossible to sample. Their probability can become sufficiently small that no sustained recovery is observed within the training horizon. The experiments that follow illustrate this behavior and examine interventions intended to preserve informative sampling and learning.

\section{Empirical Evidence of Zero Collapse}
\label{sec:empirical}
\begin{figure}[!htbp]
\centering
\includegraphics[width=\linewidth,trim=202 23 225 0,clip]{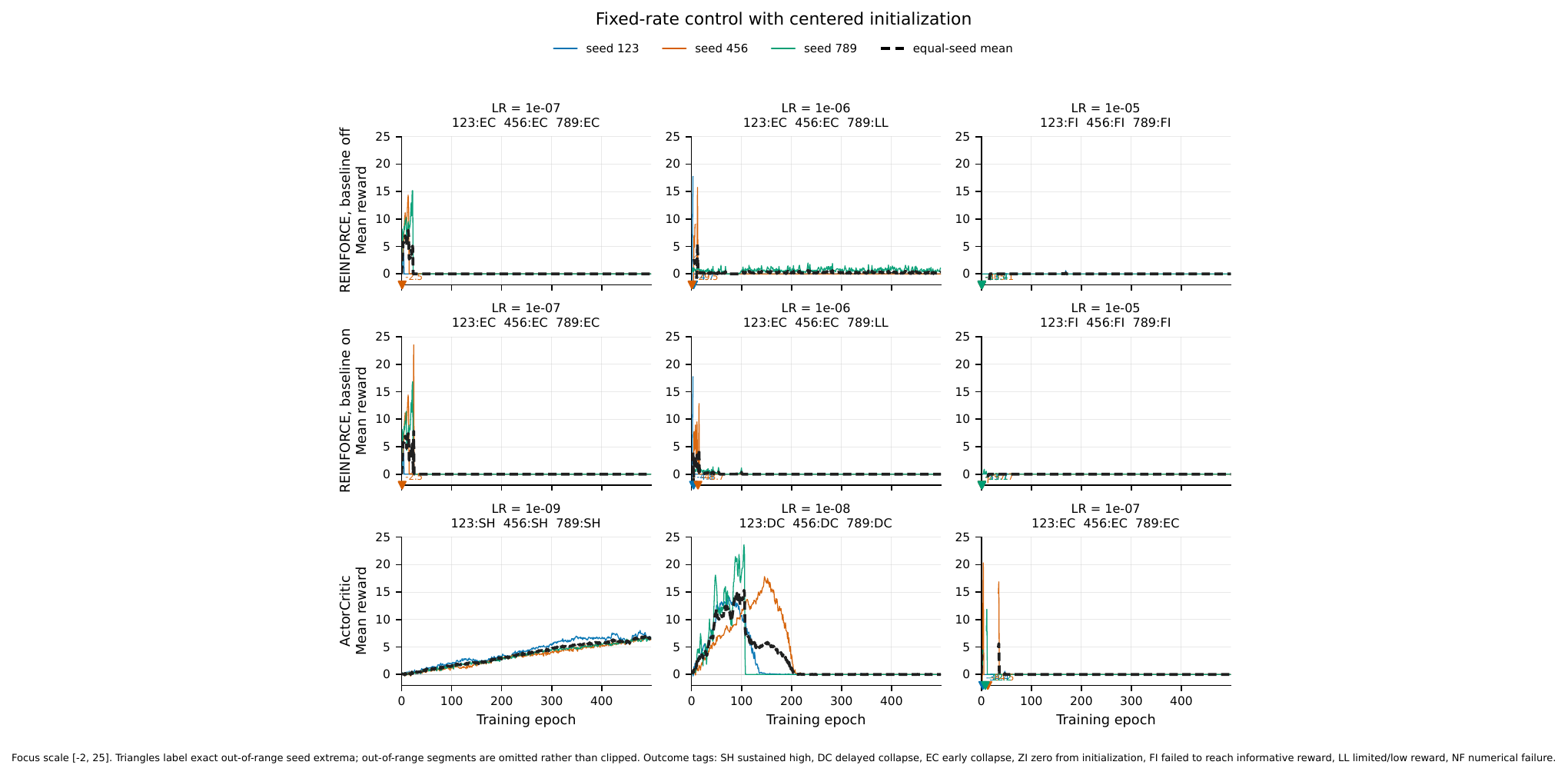}
\caption{Fixed-rate SGD controls with centered initialization and ReLU hidden/output activations. Rows show REINFORCE without a baseline, REINFORCE with a baseline, and TD state-value actor-critic; columns show algorithm-specific learning rates. Colored curves show individual seeds, and dashed black curves show their equal-weight mean over 500 epochs. The vertical axis emphasizes rewards in $[-2,25]$; out-of-range segments are omitted, with triangles and numerical labels indicating seed extrema. Panel labels identify each seed's outcome: EC, early collapse; DC, delayed collapse; FI, failed to reach informative reward; LL, limited/low reward; SH, sustained high. Operational definitions, including the threshold used for SH, appear in Appendix~\ref{sec:outcome_classification}.}
\label{fig:pre_mitigation_results}
\end{figure}

We evaluate policy-gradient methods in the simulated auction environment exhibiting the discontinuous reward structure described in Section~\ref{sec:reward_geometry}. We consider REINFORCE with and without an action-independent baseline, and actor-critic using a TD state-value critic. Figure~\ref{fig:pre_mitigation_results} presents fixed-rate SGD controls using a Gaussian policy with ReLU hidden and output activations and centered initialization. Each configuration is trained for 500 epochs using three random seeds (123, 456, and 789). Each epoch collects 50 five-step trajectories. Performance is measured by average cumulative reward, with a maximum of 25. Implementation details are given in Appendix~\ref{sec:experimental_details}.

Centered initialization sets the initial bid mean to 10 and standard deviation to 0.5, placing bids above the winning threshold of 5. This lets us study collapse after access to informative outcomes, rather than failure to obtain them from initialization (Section 6.2).

For REINFORCE, behavior depends on the learning rate. At $10^{-7}$, all three seeds show brief initial improvement followed by early collapse to near-zero reward, both with and without a baseline. At $10^{-6}$, two seeds collapse early and the third remains at low reward in each baseline condition. At $10^{-5}$, all three seeds fail to reach informative reward, rather than improving and then collapsing.

The TD state-value actor-critic provides a clear example of collapse after substantial initial learning. At a learning rate of $10^{-8}$, all three seeds improve before dropping to near-zero reward and remaining there for the rest of training. At $10^{-7}$, collapse occurs much earlier. At $10^{-9}$, learning proceeds slowly and no collapse is observed within 500 epochs; these runs do not establish whether learning would remain stable as the policy approaches higher-reward behavior.

The central observation is that initial learning---even substantial improvement---can be followed by a persistent near-zero-reward state. This behavior is consistent with the loss of informative samples described in Section~\ref{sec:mechanism}. The following sections examine interventions and a critic diagnostic to investigate the dynamics underlying this behavior.

\section{Mitigating Zero Collapse}
\label{sec:mitigation}
\subsection{Step Size / Update Magnitude Control}
\label{sec:adaptive_step}
One factor that can contribute to zero collapse is the magnitude of policy updates. Under fixed-rate SGD, the parameter displacement depends on both the learning rate and the magnitude of the gradient estimate. A small nominal learning rate therefore does not necessarily imply a small update. Near the reward discontinuity, an update that shifts the policy into the flat region can sharply reduce the frequency of informative samples, making subsequent recovery difficult.

To control update magnitude, we use SGD with an adaptive learning rate
\[
\alpha_k=\sqrt{\frac{\epsilon}{\|g_k\|_2^2+\eta}},
\]
where $\epsilon>0$ controls the update scale, $\eta=10^{-12}$ provides numerical stabilization, and $g_k$ is the gradient of the training loss over all parameters being updated. For REINFORCE without a baseline, this includes only policy parameters. For REINFORCE with a learned baseline and for the TD actor-critic, it includes both policy and value-network parameters, with the value loss weighted by $0.5$.

For SGD, this choice gives
\[
\|\Delta\theta_k\|_2^2
=\alpha_k^2\|g_k\|_2^2
=\epsilon\frac{\|g_k\|_2^2}{\|g_k\|_2^2+\eta}
\leq\epsilon.
\]
Thus, \(\epsilon\) sets an upper bound on the squared parameter displacement, with \(\|\Delta\theta_k\|_2\approx\sqrt{\epsilon}\) when \(\|g_k\|_2^2\gg\eta\). When a value network is present, this bound applies to the joint policy and value-network update; the actor’s share depends on their relative gradient magnitudes. This bound controls movement in parameter space, rather than the resulting change in the policy’s action distribution.

\begin{figure}[!htbp]
\centering
\includegraphics[width=\linewidth,trim=86 23 110 0,clip]{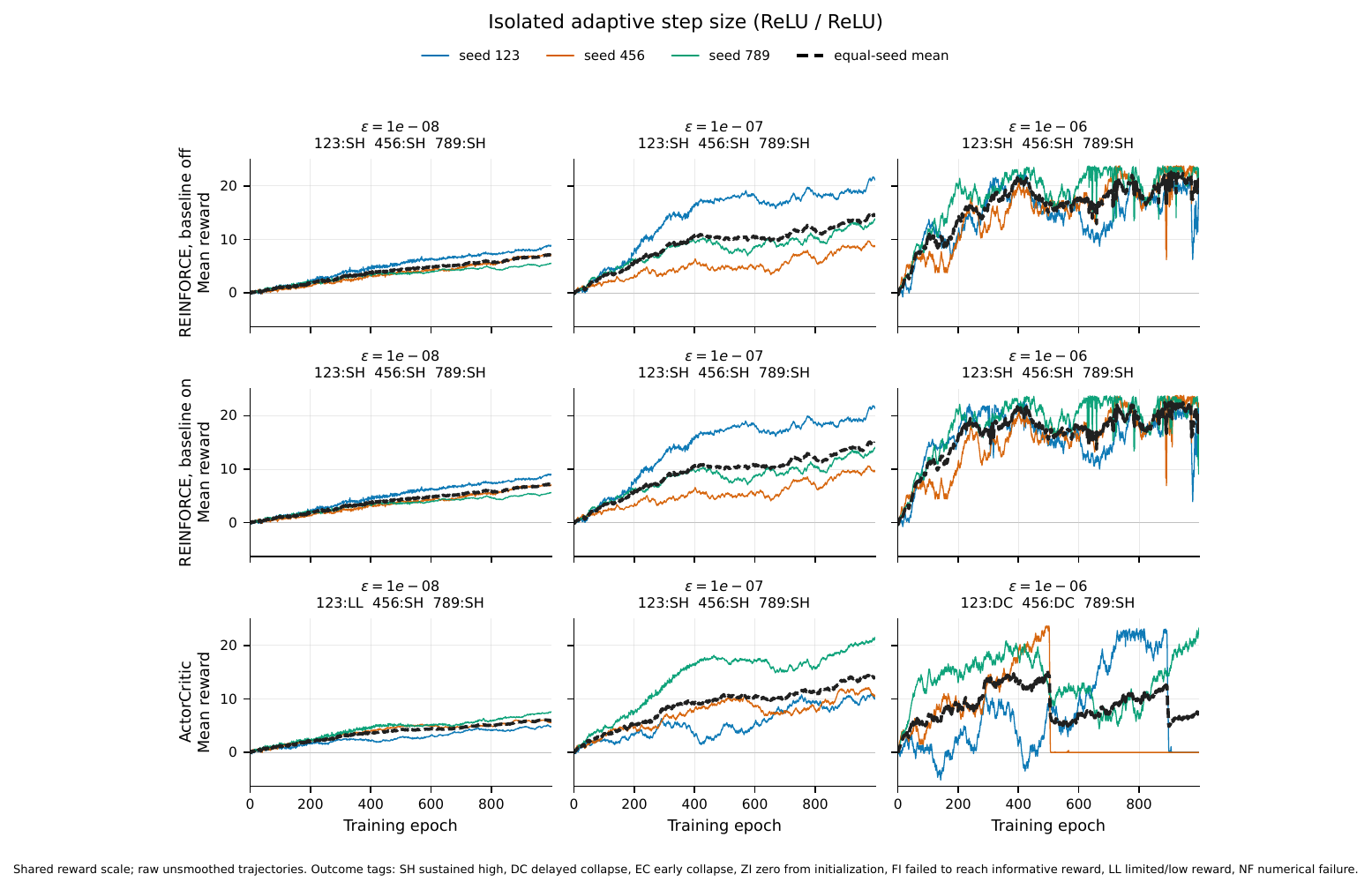}
\caption{Isolated adaptive-step intervention with centered initialization, ReLU/ReLU hidden/output activations, and SGD over 1,000 epochs. Rows distinguish REINFORCE without and with a learned baseline and the TD state-value actor-critic; columns vary $\epsilon$. Colored curves show individual seeds and dashed curves their equal-weight mean. REINFORCE sustains high reward across the tested settings. Actor-critic outcomes include sustained learning, low reward, and delayed collapse. The normalization includes value-network gradients when present. Outcome codes are defined in Appendix~\ref{sec:outcome_classification}.}
\label{fig:adaptive_step}
\end{figure}

Figure~\ref{fig:adaptive_step} isolates this intervention using the same ReLU hidden and output activations and centered initialization as the fixed-rate controls. With adaptive step-size control, REINFORCE sustains high reward across all three seeds at each tested setting, both with and without a baseline, over 1,000 epochs. The TD state-value actor-critic also achieves sustained high reward in several configurations, although some runs retain low reward or undergo delayed collapse. These results support update-magnitude control as a practical mitigation in this setting.

The normalization does not correct an inaccurate gradient direction. The following subsections consider initialization, policy parameterization, and the use of a baseline as additional factors affecting learning near the discontinuity.

\subsection{Initialization Strategy}
Policy initialization can place much of the initial action distribution within the flat, zero-reward region. When sampled trajectories yield zero return, REINFORCE receives no reward-weighted learning signal, and the policy may fail to reach informative actions in the first place. This differs from collapse after an initial period of learning, although both leave the policy with little signal for improvement.

A practical initialization strategy is therefore to place sufficient probability mass in the informative region. This need not require precise knowledge of the reward threshold. In the auction setting, a coarse estimate of the bid at which payment offsets the value of winning can provide a reference point above the threshold. Initializing the action distribution around this reference increases the chance of sampling actions with informative returns.

Table~\ref{tab:rl_experiment_outcomes} summarizes the initialization comparisons. With uncentered Glorot initialization and ReLU hidden and output activations, most runs remain at zero reward from initialization. Centering the output bias alone produces mixed outcomes, including runs that still remain at zero. The centered zero-head initialization used in Figure~\ref{fig:pre_mitigation_results} provides a more controlled starting point for studying subsequent learning dynamics. It initializes the bid distribution at mean 10 and standard deviation 0.5, independently of the state; Appendix~\ref{sec:initialization} gives the construction.

Initialization addresses access to informative samples at the start of training. The early and delayed collapses observed with centered initialization show that this alone does not prevent the policy from subsequently entering a low-signal region.

\subsection{Smooth Policy Parameterization}
Another approach to mitigating zero collapse is to change the policy parameterization, particularly the output activation. The parameterization affects both how gradients propagate through the network and how parameter updates change the action distribution.

For an output preactivation $z$, ReLU has derivative one when $z>0$ and zero when $z<0$. In contrast, softplus,
\[
\operatorname{softplus}(z)=\log(1+e^z),\qquad
\operatorname{softplus}'(z)=\frac{1}{1+e^{-z}},
\]
has a derivative that varies smoothly between zero and one. As the preactivation decreases, softplus attenuates the output response to a small change in that preactivation, while avoiding ReLU's exactly inactive negative branch. These properties offer a possible way to moderate changes in the policy outputs. The resulting change in the action distribution also depends on the rest of the network and the current policy parameters.

\begin{figure}[!htbp]
\centering
\includegraphics[width=\linewidth]{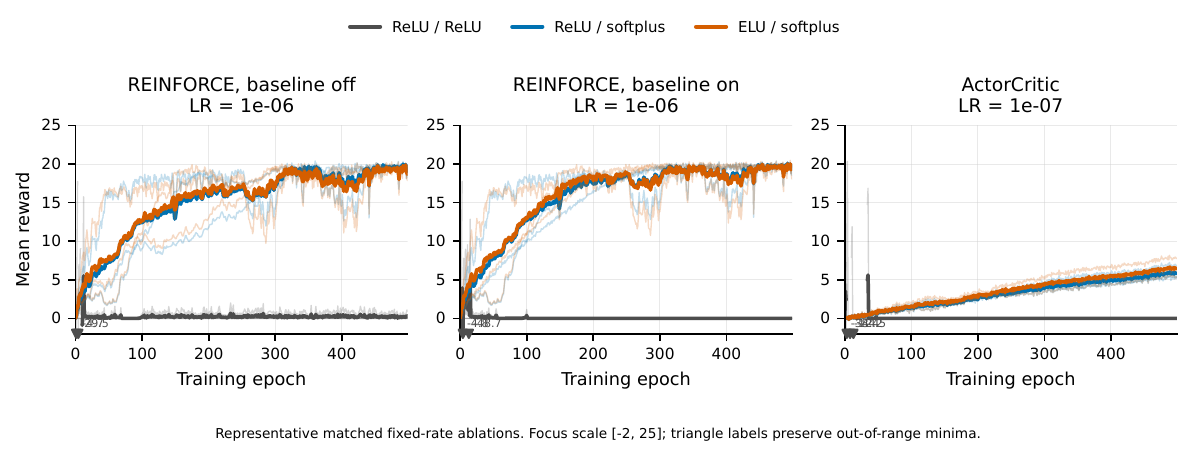}
\caption{Representative matched fixed-rate parameterization comparisons over 500 epochs. Panels show REINFORCE without a baseline, REINFORCE with a baseline (both at learning rate $10^{-6}$), and TD state-value actor-critic (at $10^{-7}$). Colors distinguish hidden/output activation pairs; faint curves show seeds and heavy curves their equal-weight means. All use centered initialization and SGD. The vertical axis emphasizes $[-2,25]$, with triangle labels indicating out-of-range minima. Figure~\ref{fig:smooth_parameterization_full} contains the full factorial.}
\label{fig:smooth_compact}
\end{figure}

Figure~\ref{fig:smooth_compact} compares ReLU hidden/output activations with ReLU/softplus and ELU/softplus parameterizations under matched fixed-rate SGD settings and centered initialization. At the representative learning rates shown---$10^{-6}$ for REINFORCE with and without a baseline, and $10^{-7}$ for the TD state-value actor-critic---both softplus-output variants sustain high reward across the three seeds, whereas the ReLU/ReLU controls collapse or remain at low reward.

The full factorial comparison in Figure~\ref{fig:smooth_parameterization_full} shows how this behavior varies across learning rates. For example, at the smaller actor-critic learning rates, the softplus-output variants make little progress within the training horizon. Thus, changing the parameterization also changes the effective learning scale. The representative comparisons establish that smooth-output parameterizations can mitigate the observed failure, while the derivative properties above provide a plausible interpretation of their effect.

\subsection{Baseline as a Stabilizing Signal}
An action-independent baseline leaves the expected REINFORCE gradient unchanged but can reduce the variability of sampled updates \cite{williams1992reinforce,sutton2000policy}. Here, its relevance is the stability of learning near the reward threshold.

Near the reward threshold, the policy may still sample both informative and zero-return outcomes. With a positive baseline, zero-return samples receive negative weights, while samples whose returns exceed the baseline receive positive weights. This allows below-baseline outcomes to contribute to the finite-sample update rather than contributing zero under raw-return weighting. The resulting stabilization can help preserve learning near the cliff or during an early stage of collapse when informative actions are still sampled. It does not provide a systematic recovery direction once informative samples are effectively absent.

The REINFORCE/SGD panels of Figure~\ref{fig:matched_fixed_optimizer} show a substantial stabilization effect from the baseline at a learning rate of $10^{-5}$, using the ReLU/softplus parameterization. Without a baseline, reward repeatedly drops below the high-reward region, producing the scattered, ``rain-like'' pattern. With a baseline, these excursions are markedly reduced and high reward is maintained more consistently. At the more aggressive rate of $10^{-4}$, the baseline does not prevent failed or numerically unstable runs.

Under adaptive step-size control, the difference between baseline-off and baseline-on behavior is much smaller. In the isolated intervention shown in Figure~\ref{fig:adaptive_step}, both conditions sustain high reward across all three seeds at every tested setting. Here the learned baseline also contributes gradients to the shared normalization, so this comparison reflects both return centering and its effect on the update scale.

\section{Mitigation Results}
\label{sec:mitigation_results}
\begin{figure}[!htbp]
\centering
\includegraphics[width=\linewidth,trim=85 23 110 0,clip]{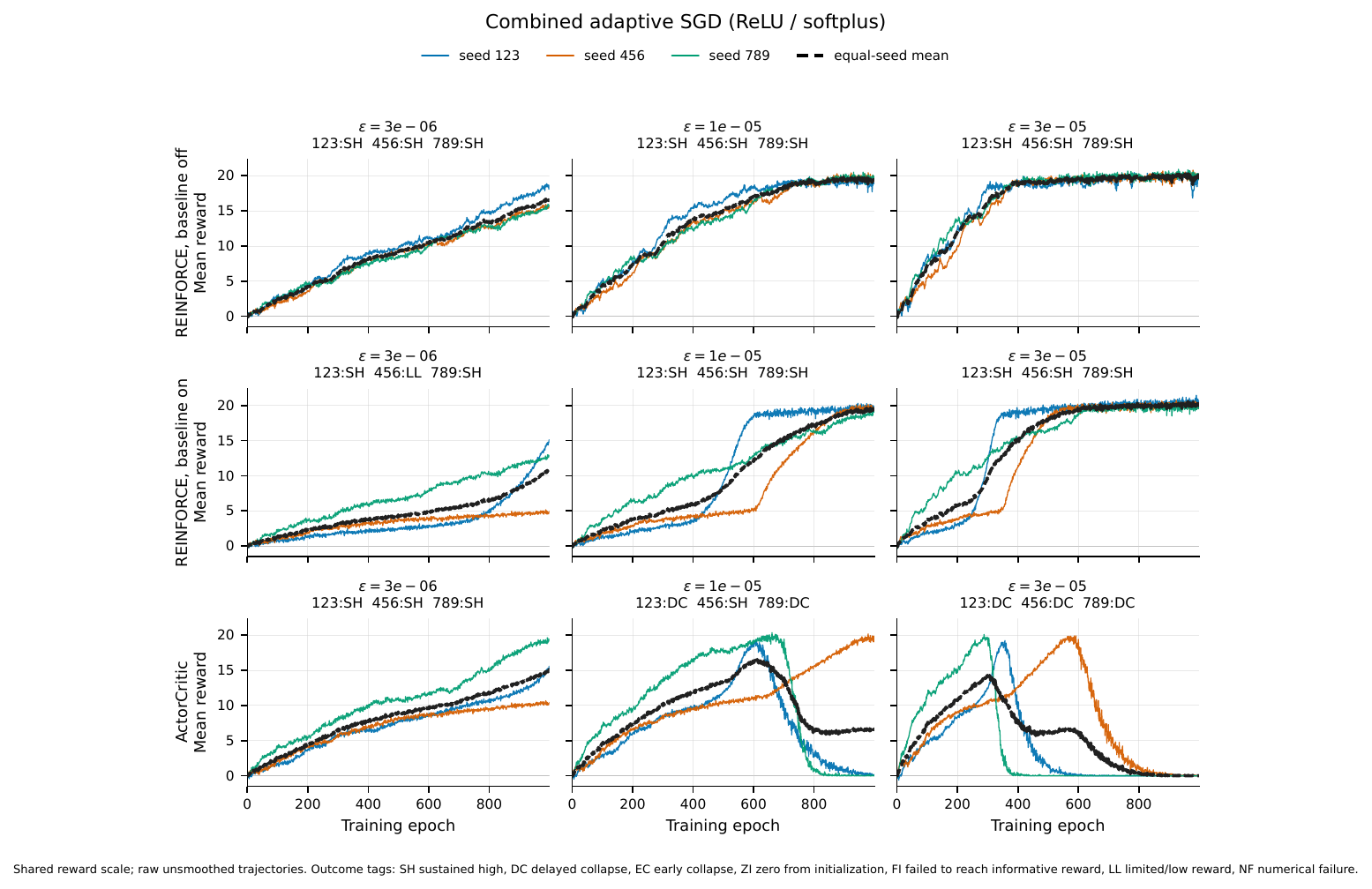}
\caption{Combined adaptive-SGD and smooth-output treatment with centered initialization. Rows show REINFORCE without a baseline, REINFORCE with a baseline, and TD state-value actor-critic, all using ReLU hidden activations and softplus outputs. Columns vary the update-scale parameter $\epsilon$. Colored curves show raw epoch-level mean rewards for individual seeds over 1,000 epochs; dashed black curves show their equal-weight mean. REINFORCE maintains high reward at the two larger settings, whereas the TD actor-critic exhibits delayed collapse in two seeds at $\epsilon=10^{-5}$ and all three seeds at $\epsilon=3\times10^{-5}$. Panel labels identify seed-level outcomes: SH, sustained high; DC, delayed collapse; LL, limited/low reward. See Appendix~\ref{sec:outcome_classification} for operational definitions.}
\label{fig:post_mitigation_results}
\end{figure}

The isolated interventions in Figures~\ref{fig:adaptive_step} and~\ref{fig:smooth_compact} show that both adaptive step-size control and smooth-output parameterization can mitigate zero collapse in this environment. We next evaluate their combined effect using adaptive SGD with ReLU hidden activations, softplus outputs, and the centered initialization used in the preceding experiments. Figure~\ref{fig:post_mitigation_results} shows three seeds per configuration over 1,000 training epochs.

For REINFORCE, the combined treatment supports sustained learning with and without a baseline. At $\epsilon=10^{-5}$ and $3\times10^{-5}$, all three seeds in both baseline conditions reach and maintain reward around 20. At $\epsilon=3\times10^{-6}$, learning is slower, particularly with a baseline, where one seed remains at low reward. None of these REINFORCE runs exhibits the persistent collapse to near-zero reward seen in the fixed-rate controls. Qualitatively, the combined treatment produces more consistent sustained learning across the tested seeds than the fixed-rate runs.

The TD state-value actor-critic retains a clear failure mode under the combined treatment. At $\epsilon=3\times10^{-6}$, all three seeds continue learning throughout the observed horizon. At $\epsilon=10^{-5}$, two seeds improve substantially before collapsing, while the third continues to improve. At $\epsilon=3\times10^{-5}$, all three seeds reach substantial positive reward before collapsing to near zero. Thus, the combined treatment supports sustained REINFORCE performance at settings where the TD actor-critic still undergoes delayed collapse.

These results demonstrate practical mitigations for the observed failure while showing that update-magnitude control and smooth outputs do not eliminate it in the TD actor-critic. The next section examines the learned critic near one observed collapse.

\section{Critic Misvaluation Near Collapse}
\label{sec:critic_mechanism}
\begin{figure}[!htbp]
\centering
\includegraphics[width=.92\linewidth]{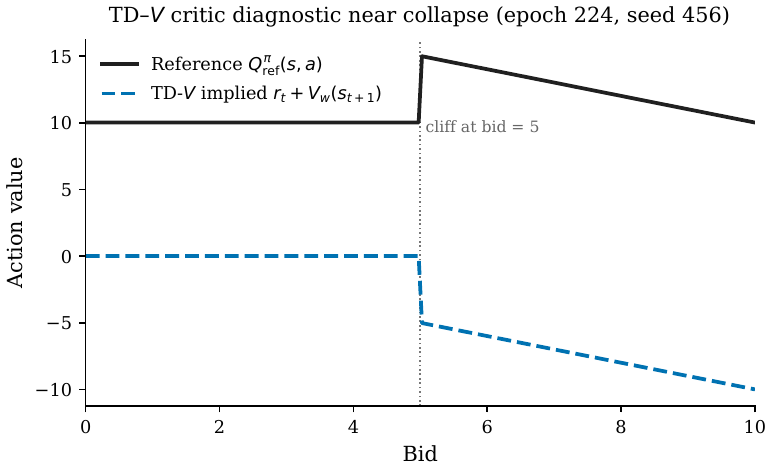}
\caption{TD state-value critic diagnostic near collapse at epoch 224, seed 456, from a diagnostic rerun of the fixed-rate SGD control at $10^{-8}$. The solid black curve shows reference action values $Q^\pi_{\mathrm{ref}}(s,a)$; the dashed blue curve shows the learned critic's implied action values $r_t+V_w(s_{t+1})$. The dotted vertical line marks the bid-5 threshold. Reference values rise across the threshold, whereas the critic-implied values fall, reversing the relative desirability of actions immediately below and above the cliff. Appendix~\ref{sec:critic_diagnostic_protocol} gives the diagnostic protocol.}
\label{fig:critic_diagnostic}
\end{figure}

The delayed collapses observed in the TD state-value actor-critic motivate examining the learning signal supplied by its critic. Figure~\ref{fig:critic_diagnostic} examines a diagnostic rerun of the fixed-rate SGD control at learning rate $10^{-8}$, seed 456, using a checkpoint at epoch 224. At a fixed state with two decisions remaining, we sweep the current bid and compare reference action values with those implied by the learned critic. The reference combines the exact immediate auction transition with a Monte Carlo estimate of the final reward under the same learned policy. The critic-implied values instead use $V_w(s_{t+1})$ to estimate continuation.

The reference action values remain positive on both sides of the bid-5 threshold. Below the threshold, they are approximately 10, reflecting the value of continuing under the policy even when the current bid does not win. Immediately above the threshold, the reference value rises to approximately 15 before decreasing with the bid. Thus, actions just above the threshold are preferable to actions below it.

The learned TD-$V$ critic implies a qualitatively different ordering. Its implied action values are approximately zero below the threshold and fall to approximately $-5$ immediately above it, decreasing further as the bid increases. The discrepancy therefore goes beyond underestimating the magnitude of returns: the critic severely misvalues continuation and reverses action desirability across the cliff.

This reversal directly affects the actor's learning signal. The TD advantage subtracts the current-state value from the implied action value,
\[
\delta_t=r_t+V_w(s_{t+1})-V_w(s_t).
\]
At a fixed state, subtracting $V_w(s_t)$ preserves the ordering across actions. Consequently, the actor receives a signal that favors bids below the threshold over bids just above it, contrary to the reference action values. Such a signal can shift probability mass toward the zero-reward region and contribute to collapse.

The diagnostic therefore identifies a concrete critic error consistent with the observed failure in this run. The implied action-value curve retains a discontinuity, but its jump has the opposite sign from the reference. This also explains why limiting update magnitude alone may be insufficient: smaller updates can still move the policy toward the flat region when the critic supplies an incorrect relative valuation of actions.

\section{Discussion and Future Work}
The experiments illustrate how learning can deteriorate after initial improvement in this discontinuous-reward setting. Adaptive update-magnitude control and smooth-output parameterization support sustained learning in configurations where the fixed-rate controls collapse. The baseline comparison further shows that centering returns can substantially stabilize REINFORCE when update magnitude is otherwise uncontrolled.

The matched SGD--Adam comparison in Figure~\ref{fig:matched_fixed_optimizer} shows varied behavior across the tested Adam configurations: sustained learning, large reward fluctuations with recovery in some runs, and persistent low reward or failure in others.

For the TD state-value actor-critic, Figure~\ref{fig:critic_diagnostic} highlights the consequences of acting on valuations before they are sufficiently accurate. The critic's continuation estimates reverse the relative desirability of actions across the reward threshold, supplying the actor with a misleading learning signal. Controlling update magnitude can moderate the effect of such errors, but cannot correct the underlying action ranking. This motivates investigating how policy improvement can account for the reliability of the critic's estimates, particularly near the threshold.

A complementary direction is to examine methods that control policy change more directly, such as TRPO \cite{schulman2015trpo} and PPO \cite{schulman2017ppo}. Our adaptive rule bounds parameter displacement, whereas the resulting change in the action distribution depends on the policy parameterization. Evaluating whether policy-space update control helps preserve informative sampling would clarify its role in mitigating zero collapse.

The auction setting provides a concrete demonstration of zero collapse and several practical mitigations. An open question is how these dynamics extend to other environments where a reward threshold separates informative actions from a flat, low-signal region.

\section{Conclusion}
We identify and illustrate zero collapse in a repeated first-price auction setting: a policy initially improves, then shifts into persistently near-zero reward as informative outcomes become rarely sampled. The threshold separating winning and losing bids helps explain how policy updates can move learning into a region that provides little signal for recovery.

The experiments demonstrate several practical mitigations. Centered initialization provides access to informative outcomes at the start of training, while adaptive update-magnitude control and smooth-output parameterization support sustained learning in configurations where fixed-rate controls collapse. A learned baseline substantially stabilizes REINFORCE in the fixed-rate comparison, with a smaller observed difference under adaptive control.

The TD state-value actor-critic still exhibits delayed collapse under some mitigation settings. A diagnostic near one observed collapse reveals severe continuation-value errors that reverse action desirability across the winning threshold. This finding highlights the consequences of an actor relying on valuations before they are sufficiently accurate.

Together, these results show how thresholded auction outcomes, policy updates, and value estimation interact to produce zero collapse, and demonstrate practical ways to mitigate it in the studied setting.

\clearpage
\appendix
\section{Environment and Problem Formulation}

\subsection{Environment: Ad Exchange Simulation}

We consider a stylized ad exchange (AdX) environment in which multiple agents compete in repeated first-price auctions over a finite horizon.

Each agent is assigned a single advertising campaign and participates in a sequence of auctions for impression opportunities (i.e., opportunities to display ads to users). The agent’s objective is to maximize profit, defined as total value obtained from fulfilling the campaign minus total spend incurred through bidding.

A campaign is defined by:

\begin{itemize}
    \item a \textbf{market segment} $s_{\text{seg}}$, corresponding to a subset of user attributes (e.g., demographic features)
    \item a \textbf{reach} $R$, specifying the required number of impressions
    \item a \textbf{budget} $B$, representing the reward received upon successful completion
\end{itemize}

An example campaign is:
\[
( \text{segment} = \text{Female\_Old}, \quad R = 500, \quad B = 40 )
\]

To fulfill a campaign, an agent must obtain at least $R$ impressions from users matching the specified segment. This requires repeatedly winning auctions for relevant users. Winning an auction incurs cost, while the campaign reward is realized proportional to the reach requirement by the end of the episode.

\subsection{Markov Game Formulation}

We model the environment as a fully observable Markov Game:
\[
\langle \mathcal{S}, \mathcal{A}, \mathcal{R}, \mathcal{T}, \gamma \rangle.
\]

Let $m$ denote the number of agents.

\subsubsection{State}

The global state is a product of agent-specific states:
\[
\mathcal{S} = \prod_{i=1}^m \mathcal{S}^{(i)}.
\]

Each agent state $s^{(i)} \in \mathcal{S}^{(i)}$ includes:
\begin{itemize}
    \item the assigned campaign
    \item cumulative spend
    \item number of impressions obtained
    \item the current time step $t$
\end{itemize}

\subsubsection{Actions}

At each time step $t$, each agent $i$ selects an action:
\[
a_t^{(i)} \in \mathcal{A}^{(i)},
\]
where the action specifies bids across market segments:
\[
a_t^{(i)} = \{(b_{t,1}, u_{t,1}), \dots, (b_{t,K}, u_{t,K})\},
\]

with $b_{t, j}$ denoting the bid for segment $j$ and $u_{t, j}$ the allocated budget.

\subsubsection{Rewards}

The reward for agent $i$ at time $t$ is given by:
\[
r^{(i)}(s_t, a_t^{(1)}, \dots, a_t^{(m)}) = \text{value} - \text{cost}.
\]

Agents incur expenditure throughout the episode. At termination, agent $i$ receives campaign value $\min(B_i,I_iB_i/R_i)$, where $I_i$ is its final number of impressions. Partial fulfillment therefore earns proportional value; reaching $R_i$ caps that value rather than switching it on.

\subsection{Single-Agent Reduction}

To analyze learning dynamics, we consider a single learning agent $i$, while all other agents follow fixed policies.

This induces a Markov Decision Process (MDP) \cite{sutton2018rl}:
\[
\langle \mathcal{S}, \mathcal{A}, \mathcal{R}, \mathcal{T}, \gamma \rangle,
\]
where the action space corresponds to that of agent $i$, i.e., $a_t = a_t^{(i)}$, and the dynamics incorporate the behavior of the other agents through their fixed policies.

The reward function is given by:
\[
r(s_t, a_t) = r^{(i)}(s_t, a_t^{(1)}, \dots, a_t^{(m)}),
\]
where $a_t^{(j)} \sim \pi_{\theta_j}(s_t)$ for $j \neq i$.

Similarly, the transition dynamics are defined as:
\[
\mathcal{T}(s_{t+1} \mid s_t, a_t)
= \mathcal{T}(s_{t+1} \mid s_t, a_t^{(1)}, \dots, a_t^{(m)}),
\]
with other agents’ actions sampled from their fixed policies.

This reduction allows us to analyze policy gradient methods in a standard single-agent setting, while preserving the stochasticity induced by multi-agent interaction.

\subsection{Structural Properties of the Reward}
Auction allocation changes discontinuously when a bid crosses the competing bid. A win incurs immediate expenditure and changes the impression count used to calculate terminal campaign value. The value component is delayed until termination and is proportional to fulfillment up to the campaign cap; it does not require full reach to become positive.

A policy that loses every auction incurs no expenditure and earns no campaign value, producing an all-zero trajectory. This connects the repeated-game setting to the simplified payoff geometry of Section~\ref{sec:reward_geometry}. Losing one auction, however, need not imply zero return because later auctions can still contribute campaign value.

\section{Multi-Agent Policy Factorization}
\label{sec:factorization}

We assume that agents act independently and simultaneously conditioned on the state.

\paragraph{Assumption (Conditional Independence).}
The joint policy factorizes across agents:
\[
\pi_\theta(a_t^{(1)}, \dots, a_t^{(m)} \mid s_t)
= \prod_{i=1}^m \pi_{\theta_i}(a_t^{(i)} \mid s_t).
\]

\paragraph{Trajectory Distribution.}
Under this assumption, the trajectory distribution can be written as:
\[
p_\theta(\tau)
= p(s_0) \prod_{t=0}^{T-1}
\left[
\left( \prod_{i=1}^m \pi_{\theta_i}(a_t^{(i)} \mid s_t) \right)
\mathcal{T}(s_{t+1} \mid s_t, a_t^{(1)}, \dots, a_t^{(m)})
\right].
\]

\paragraph{Implication for Gradients.}
The factorization of the joint policy implies that the policy-dependent factor in the trajectory likelihood has log-probability:
\[
\log \pi_\theta(\tau)
= \sum_{t=0}^{T-1} \sum_{i=1}^m
\log \pi_{\theta_i}(a_t^{(i)} \mid s_t).
\]

As a result, gradients with respect to each agent's parameters depend only on its own policy:
\[
\nabla_{\theta_i} \log \pi_\theta(\tau)
= \sum_{t=0}^{T-1}
\nabla_{\theta_i} \log \pi_{\theta_i}(a_t^{(i)} \mid s_t).
\]

This decomposition enables per-agent policy gradient updates while treating other agents as part of the environment.

\section{Policy Gradient Derivations}

\subsection{Value Functions and Bellman Relations}

For agent $i$, define the value functions:
\[
V^{\pi_i}(s_t) = \mathbb{E}\left[\sum_{t'=t}^{T-1} r^{(i)}_{t'} \mid s_t \right],
\quad
Q^{\pi_i}(s_t, \mathbf{a}_t) = \mathbb{E}\left[\sum_{t'=t}^{T-1} r^{(i)}_{t'} \mid s_t, \mathbf{a}_t \right].
\]

These satisfy the Bellman relation:
\[
Q^{\pi_i}(s_t, \mathbf{a}_t)
= r^{(i)}_t + \mathbb{E}_{s_{t+1}}\left[V^{\pi_i}(s_{t+1})\right],
\]
Here $\mathbf{a}_t=(a_t^{(1)},\ldots,a_t^{(m)})$ denotes the joint action. The expectation is over the next-state distribution, and $r_t^{(i)}$ is the expected immediate reward conditional on the current state and joint action.

\subsection{Multi-Agent Policy Gradient}

The objective for agent $i$ is:
\[
J_i(\theta_i) = \mathbb{E}_{\tau \sim p_\theta(\tau)}\left[r^{(i)}(\tau)\right].
\]

Using the log-derivative trick and the factorization from Appendix~\ref{sec:factorization}:
\[
\nabla_{\theta_i} J_i
= \mathbb{E}_{\tau \sim p_\theta(\tau)}\left[
\sum_{t=0}^{T-1}
\nabla_{\theta_i} \log \pi_{\theta_i}(a_t^{(i)} \mid s_t)
\left(\sum_{t'=t}^{T-1} r^{(i)}_{t'}\right)
\right].
\]

This yields the standard REINFORCE estimator:
\[
\nabla_{\theta_i} J_i
= \mathbb{E}_{\tau \sim p_\theta(\tau)}\left[
\sum_{t}
\nabla_{\theta_i} \log \pi_{\theta_i}(a_t^{(i)} \mid s_t)
\hat{Q}^{(i)}_t
\right].
\]

Where $\hat{Q}^{(i)}_t := \sum_{t'=t}^{T-1} r^{(i)}_{t'}$.

\subsection{Baseline and Advantage Estimation}

Subtracting a baseline does not change the expectation:
\[
\nabla_{\theta_i} J_i
= \mathbb{E}_{\tau \sim p_\theta(\tau)}\left[
\sum_{t}
\nabla_{\theta_i} \log \pi_{\theta_i}(a_t^{(i)} \mid s_t)
\left(\hat{Q}^{(i)}_t - V^{\pi_i}(s_t)\right)
\right].
\]

This defines the advantage:
\[
A^{(i)}(s_t, \mathbf{a}_t)
= Q^{\pi_i}(s_t, \mathbf{a}_t) - V^{\pi_i}(s_t).
\]

\subsection{Actor-Critic Approximations}

In practice, $Q^{\pi_i}$ and $V^{\pi_i}$ are approximated with function approximators:
\[
\hat{V}_{\mathbf{w}_i}(s), \quad \hat{Q}_{\mathbf{u}_i}(s, \mathbf{a}).
\]

This yields the estimator:
\[
\nabla_{\theta_i} J_i
\approx
\mathbb{E}\left[
\sum_{t}
\nabla_{\theta_i} \log \pi_{\theta_i}(a_t^{(i)} \mid s_t)
\left(\hat{Q}_{\mathbf{u}_i}(s_t, \mathbf{a}_t) - \hat{V}_{\mathbf{w}_i}(s_t)\right)
\right].
\]

\subsection{Temporal-Difference Learning}

Using the Bellman relation, we obtain the TD advantage:
\[
A_t^{(i)}
= r^{(i)}_t + \hat{V}_{\mathbf{w}_i}(s_{t+1}) - \hat{V}_{\mathbf{w}_i}(s_t).
\]

This yields the TD actor-critic update:
\[
\nabla_{\theta_i} J_i
\approx
\mathbb{E}\left[
\sum_{t}
\nabla_{\theta_i} \log \pi_{\theta_i}(a_t^{(i)} \mid s_t)
A_t^{(i)}
\right].
\]

\paragraph{SARSA-style Advantage (On-Policy Q).}

For completeness, an alternative on-policy estimator, not evaluated in the revised experiments, replaces the bootstrap target with the value of the next sampled action:
\[
A_t^{(i)} = r^{(i)}_t + \hat{Q}_{\mathbf{u}_i}(s_{t+1}, \mathbf{a}_{t+1}) - \hat{V}_{\mathbf{w}_i}(s_t),
\]
where $a_{t+1}^{(i)} \sim \pi_{\theta_i}(\cdot \mid s_{t+1})$.

This corresponds to a SARSA-style update, in contrast to the TD estimator which uses $\hat{V}_{\mathbf{w}_i}(s_{t+1})$.

\section{Experimental Details}
\label{sec:experimental_details}
\subsection{Implementation Details}
The experiments use the shared setup in Table~\ref{tab:setup}. Learning-rate grids, activation choices, and training horizons are specified by experiment family below. Policies parameterize Gaussian action distributions. Training curves report epoch-level mean cumulative reward; where an aggregate curve is shown, seeds receive equal weight. Table~\ref{tab:rl_experiment_outcomes} records the settings, training horizons, and qualitative outcomes for individual seeds.

\begingroup
\small
\renewcommand{\arraystretch}{1.15}
\begin{longtable}{@{}p{.21\linewidth}p{.75\linewidth}@{}}
\caption{Shared implementation and training settings.}\label{tab:setup}\\
\toprule Component & Specification \\ \midrule
\endfirsthead
\toprule Component & Specification \\ \midrule
\endhead
\bottomrule\endfoot
Environment & Five-step episodes with one first-price auction per step; one learner competes against a fixed bidder with bid 5 and total limit 5.\\
Reward & Negative expenditure at each step, with capped campaign value $\min(B,IB/R)$ added at termination, where $B$ is campaign budget, $I$ is impressions obtained, and $R$ is target reach. Rewards are unscaled.\\
Sampling and reporting & Each epoch collects 50 fresh trajectories, totaling 250 transitions, followed by one full-batch update. Reported reward is the mean cumulative episode reward across these trajectories. The discount factor is $\gamma=1$.\\
Actor input & A 12-dimensional joint state comprising reach, budget, target identifier, expenditure, impressions, and timestep for each bidder.\\
Actor network & Four hidden layers of width four. Hidden/output activations vary by experiment. Gaussian scales use the output activation plus a minimum standard deviation corresponding to 0.1 in natural action units.\\
Action construction & The policy samples bid and multiplier channels. Samples are clipped to $[0,10^{15}]$ in working coordinates; one is added to the multiplier, and the total limit is constructed as bid times multiplier.\\
Scaling & State inputs are MinMax-scaled using 10,000 environment samples collected at run construction. Action scaling is derived from the sampled campaign-budget range.\\
TD critic & Four ReLU hidden layers of width four and a scalar state-value output. Targets are $r_t+\gamma V_w(s_{t+1})$, with detached bootstrap values and zero terminal bootstrap.\\
REINFORCE baseline & A learned state-value network fitted to Monte Carlo returns from the same rollout batch. Actor weights are $G_t-V_w(s_t)$; baseline-free REINFORCE uses $G_t$.\\
Optimization & Losses are summed over the rollout batch and timesteps. When a value network is present, the loss is $L_{\mathrm{policy}}+0.5L_{\mathrm{value}}$, optimized jointly with one learning rate and one optimizer step. Value losses are summed squared errors. Actor advantages are detached.\\
Optimizer configuration & SGD without momentum or Adam, using TensorFlow 2.16.1 defaults except for the specified learning rate. No gradient-norm clipping is applied.\\
Seeds & 123, 456, and 789 for each configuration.\\
\end{longtable}
\endgroup

\subsection{Centered Initialization}
\label{sec:initialization}
Centered zero-head initialization sets the Gaussian parameter-head weight matrix to zero and selects the primary bid-channel biases to produce an initial mean of 10 and standard deviation of 0.5 in natural action units, before sample clipping. The bid distribution is therefore initially independent of the state. Biases account for the output activation and the minimum scale: ReLU uses the target preactivation directly, whereas softplus uses its inverse. The multiplier channel is left at its default neutral initialization.

The bias-centered Glorot comparison uses the same bid-channel bias construction but retains Glorot-initialized head weights. Its initial outputs therefore remain state-dependent. Uncentered Glorot initialization uses zero head biases without prescribing an initial bid mean or scale.

\subsection{Adaptive Updates}
Adaptive SGD uses
\[
\alpha_k=\sqrt{\frac{\epsilon}{\|g_k\|_2^2+10^{-12}}},
\]
where the norm includes all gradients of the training loss over the parameters being updated. For baseline-free REINFORCE, these are policy gradients only. For baseline-on REINFORCE and the TD actor-critic, the norm jointly includes policy and value-network gradients, with the value-loss contribution weighted by $0.5$. The rule bounds the joint SGD parameter displacement by $\sqrt{\epsilon}$; the actor's share depends on the relative gradient magnitudes.

\subsection{Experiment Families}
\paragraph{Initialization and parameterization.}
The initialization comparisons include uncentered Glorot initialization, Glorot initialization with a centered output bias, and centered zero-head initialization. The fixed-rate controls and subsequent intervention experiments use the centered zero-head initialization. Activation pairs denote hidden-layer and output-layer activations, respectively. The parameterization comparison evaluates ReLU/ReLU, ReLU/softplus, and ELU/softplus.

\begin{figure}[H]
\centering
\includegraphics[width=\linewidth]{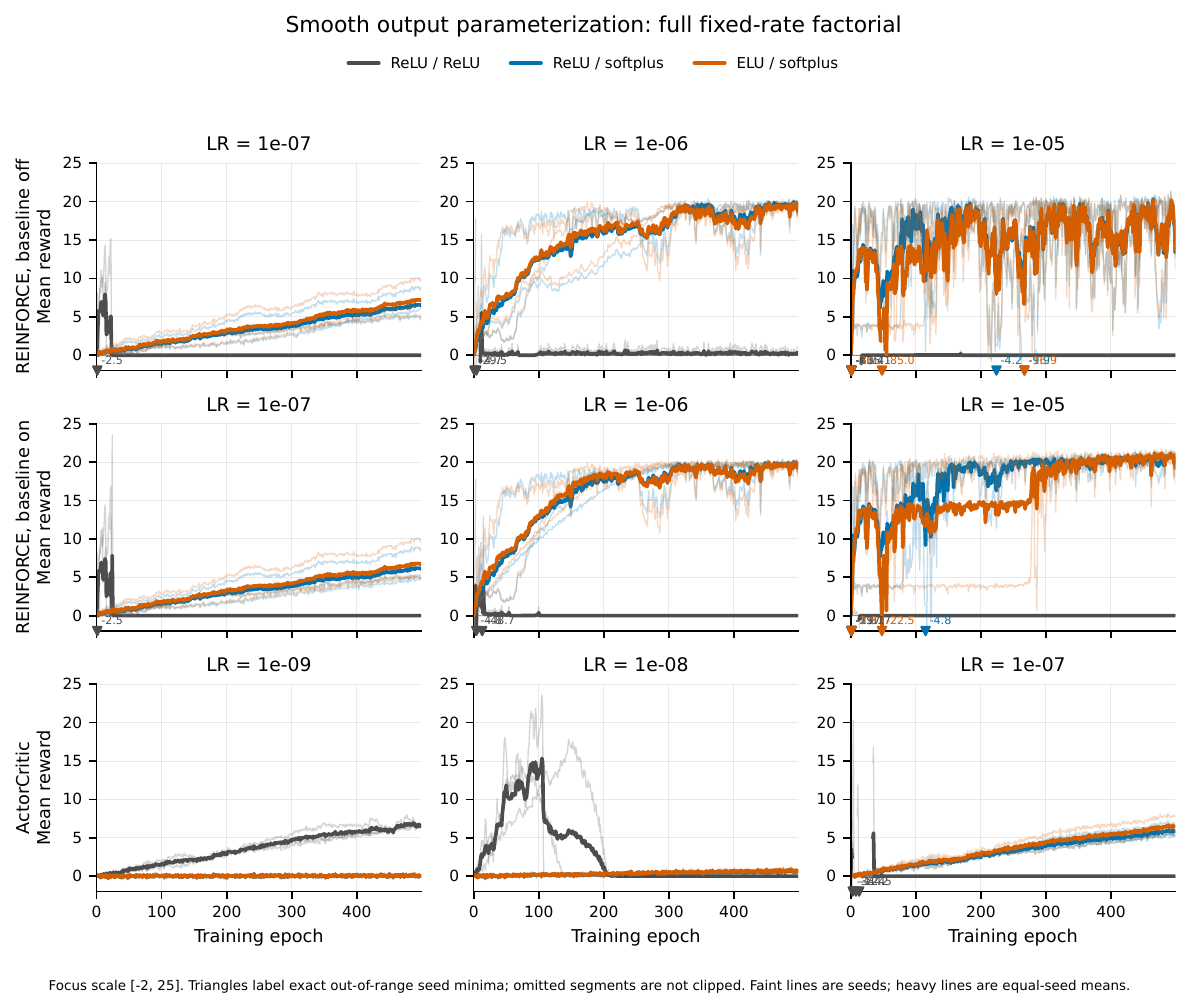}
\caption{Full fixed-rate parameterization comparison over 500 epochs, using centered initialization and SGD. Rows show REINFORCE without a baseline, REINFORCE with a baseline, and TD state-value actor-critic; columns show algorithm-specific learning rates. Colors distinguish ReLU/ReLU, ReLU/softplus, and ELU/softplus hidden/output activations. Faint curves show individual seeds and heavy curves their equal-weight mean. The ReLU/ReLU curves provide the controls shown in Figure~\ref{fig:pre_mitigation_results}; Figure~\ref{fig:smooth_compact} presents selected panels from this comparison. The vertical axis emphasizes rewards in $[-2,25]$; out-of-range segments are omitted, with triangles and numerical labels indicating seed minima.}
\label{fig:smooth_parameterization_full}
\end{figure}

\paragraph{Fixed-rate controls and parameterization ablation.}
The controls in Figure~\ref{fig:pre_mitigation_results} use SGD with ReLU/ReLU activations for 500 epochs. REINFORCE is evaluated at learning rates $10^{-7}$, $10^{-6}$, and $10^{-5}$, both with and without a baseline. The TD actor-critic is evaluated at $10^{-9}$, $10^{-8}$, and $10^{-7}$. The smooth-parameterization ablation uses the same algorithm-specific learning-rate grids and training horizon for each activation pair. Figure~\ref{fig:smooth_compact} presents representative comparisons; Figure~\ref{fig:smooth_parameterization_full} presents the complete grid.

\paragraph{Adaptive step-size experiments.}
The isolated adaptive-step intervention in Figure~\ref{fig:adaptive_step} retains ReLU/ReLU activations and evaluates $\epsilon\in\{10^{-8},10^{-7},10^{-6}\}$. The combined treatment in Figure~\ref{fig:post_mitigation_results} uses ReLU/softplus activations and evaluates $\epsilon\in\{3\times10^{-6},10^{-5},3\times10^{-5}\}$. Both families use adaptive SGD as described in Section~\ref{sec:adaptive_step} and run for 1,000 epochs. Their update-scale grids differ, so they examine each treatment across its tested settings rather than compare the treatments at identical values of $\epsilon$.

\begin{figure}[!htbp]
\centering
\includegraphics[width=\linewidth]{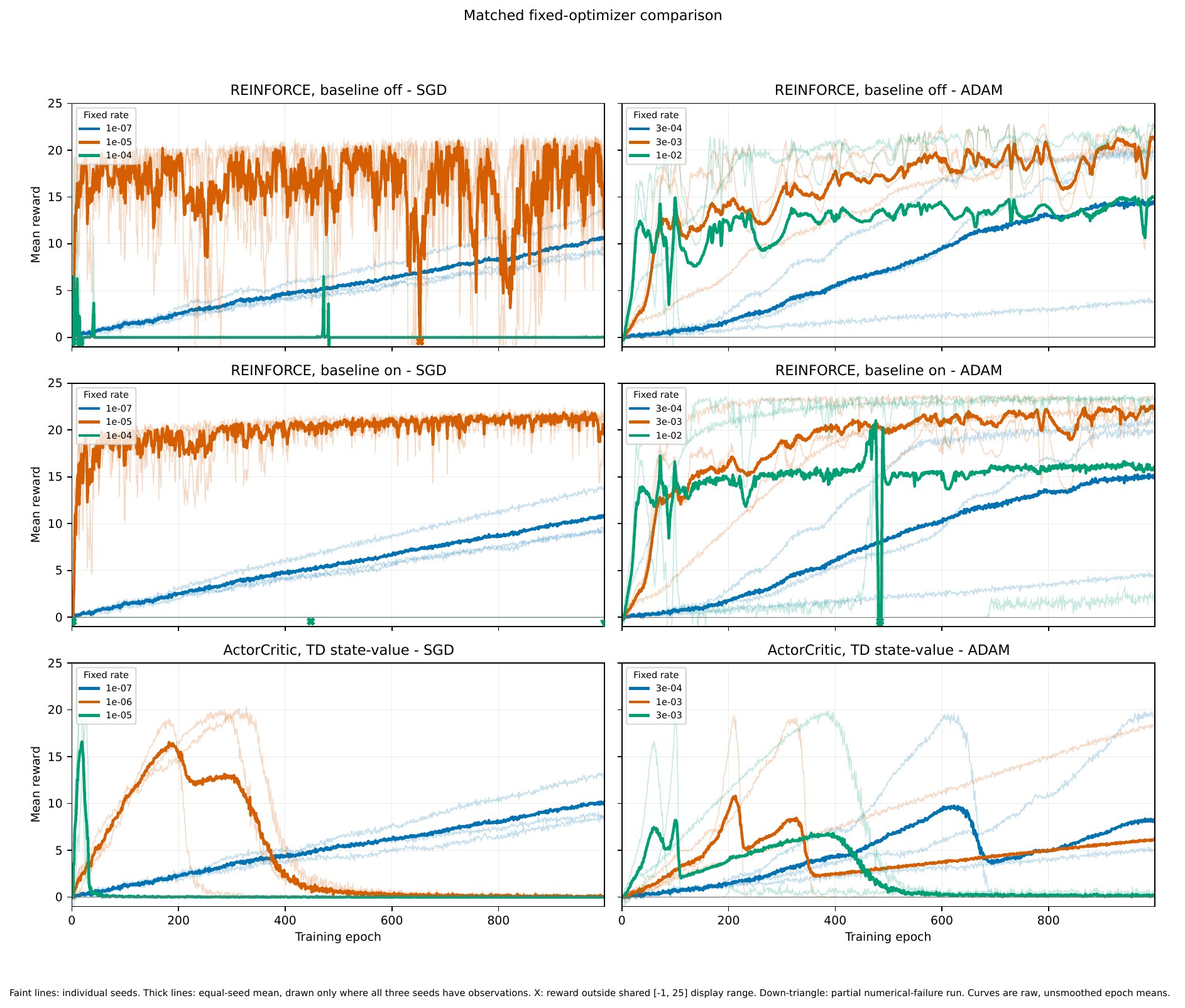}
\caption{Fixed-rate SGD and Adam comparison over a common 1,000-epoch horizon, using centered initialization and ReLU/softplus hidden/output activations. Rows show REINFORCE without a baseline, REINFORCE with a baseline, and TD state-value actor-critic; columns distinguish optimizers. Colors identify the learning rates listed in each panel. Learning-rate grids are optimizer-specific and do not represent equivalent update magnitudes. Faint curves show individual seeds; heavy curves show the equal-weight mean where all three seeds have observations. Curves are unsmoothed epoch-level mean rewards. Crosses mark rewards outside the displayed range $[-1,25]$, and downward triangles identify partial runs classified as numerical failures.}
\label{fig:matched_fixed_optimizer}
\end{figure}

\paragraph{Matched optimizer comparison.}
Figure~\ref{fig:matched_fixed_optimizer} compares fixed-rate SGD and Adam using ReLU/softplus activations, centered initialization, and a common 1,000-epoch horizon. For REINFORCE, with and without a baseline, the SGD learning rates are $\{10^{-7},10^{-5},10^{-4}\}$, and the Adam rates are $\{3\times10^{-4},3\times10^{-3},10^{-2}\}$. For the TD actor-critic, the SGD rates are $\{10^{-7},10^{-6},10^{-5}\}$, and the Adam rates are $\{3\times10^{-4},10^{-3},3\times10^{-3}\}$. These are optimizer-specific learning-rate grids; nominal rates do not represent equivalent parameter-update magnitudes. The figure displays available observations from runs that terminate with numerical failure and draws the equal-seed mean only where all three seeds have observations.

\subsection{Outcome Classification}
\label{sec:outcome_classification}
The seed-level outcome table summarizes reward traces using the ordered rules in Table~\ref{tab:outcome_rules}. Let $p$ denote the maximum epoch reward and $m$ the mean reward over the final 100 epochs.
\begin{table}[htbp]
\centering\small
\begin{tabularx}{\linewidth}{@{}p{.26\linewidth}X@{}}
\toprule Label & Operational criterion \\ \midrule
NF: numerical failure & Incomplete or nonfinite trace, or an absolute reward exceeding $10^6$.\\
ZI: zero from initialization & All epoch rewards are zero.\\
FI: failed to reach informative reward & $p<1$.\\
DC/EC: delayed/early collapse & $p\geq5$ and $m<1$. Collapse is delayed if the last epoch with reward at least 5 occurs at or after 20\% of the training horizon; otherwise it is early.\\
SH: sustained high & $m\geq5$.\\
LL: limited/low reward & Remaining runs.\\ \bottomrule
\end{tabularx}
\caption{Ordered operational rules for classifying individual reward traces.}
\label{tab:outcome_rules}
\end{table}

These labels summarize behavior within the observed training horizon. In particular, SH denotes a final-window mean of at least 5, rather than near-optimal performance or an established plateau; slowly improving runs can receive this label. The collapse labels use a near-zero final-window threshold rather than requiring exactly zero reward.

\subsection{TD State-Value Critic Diagnostic}
\label{sec:critic_diagnostic_protocol}
Figure~\ref{fig:critic_diagnostic} uses a diagnostic rerun of the centered-initialization, ReLU/ReLU actor-critic control with fixed-rate SGD, learning rate $10^{-8}$, and seed 456. The checkpoint is taken after 225 updates, corresponding to zero-based epoch 224. It was selected near the retained run's last nonzero-reward epoch, 247. The diagnostic rerun is not an exact replay of the retained training trace.

We evaluate a fixed joint state at timestep 3, with two auction decisions remaining. The unscaled state is given in Table~\ref{tab:diagnostic_state}.
\begin{table}[htbp]
\centering
\begin{tabular}{lrr}
\toprule State field & Learner & Competitor\\\midrule
Target reach & 10 & 10\\
Budget & 100 & 100\\
Target identifier & 2 & 2\\
Expenditure & 5 & 10\\
Impressions & 1 & 2\\
Timestep & 3 & 3\\\bottomrule
\end{tabular}
\caption{Fixed unscaled state for the critic diagnostic.}
\label{tab:diagnostic_state}
\end{table}

The learner's current bid is swept from 0 to 10 while its current total limit is fixed at 100. The competitor uses a fixed bid of 5 and total limit of 5. At a bid of exactly 5, allocation is equally likely between the tied bidders.

The reference evaluates the swept action followed by the checkpoint's learned policy:
\[
Q^\pi_{\mathrm{ref}}(s,a)=\mathbb{E}\left[r_t+r_{t+1}\mid s_t=s,\ a_t=a,\ \text{then follow }\pi\right].
\]
The current auction transition and immediate reward are calculated directly from the environment rules. For each resulting next state, 50,000 final actions are sampled from the learned policy and evaluated through the terminal auction and reward rules. Their mean estimates the continuation return. The discount factor is one.

The critic-implied action value uses the same current transition but substitutes the learned continuation estimate:
\[
\widehat Q_{\mathrm{TD}}(s,a)=\mathbb{E}\left[r_t+V_w(s_{t+1})\mid s_t=s,\ a_t=a\right].
\]
Expectations over tied allocations account for both possible next states. Terminal bootstrap values are zero. Both curves therefore evaluate continuation under the same learned policy; the reference is not an optimal-policy benchmark.

The estimated continuation return is approximately 10.013 after losing the current auction and 20.012--20.014 for the reported winning branches. The corresponding Monte Carlo standard errors are approximately $0.0011$, much smaller than the TD--reference discrepancies in Figure~\ref{fig:critic_diagnostic}. For comparison, the learned continuation value after losing is approximately $0.00085$.

\clearpage
\subsection{Supporting Figures and Seed-Level Outcomes}
\clearpage
\input{experiment_outcomes}

\clearpage
\section{\redact{Derby} Framework}

    Source: \texttt{https://github.com/nish-ku-121/derby}.
    
    Code and configurations are available on the \texttt{main} branch.

    \redact{Derby} is a simple bidding, auction, and market framework for creating and running auction or market games. Environments in \redact{Derby} can be interfaced in a similar fashion as environments in OpenAI's gym:
    \begin{lstlisting}[language=Python]
    env = ...
    agents = ...
    env.init(agents, num_of_days)
    for i in range(num_of_trajs):
            all_agents_states = env.reset()
            for j in range(horizon_cutoff):
                actions = []
                for agent in env.agents:
                    agent_states = env.get_folded_states(
                                        agent, all_agents_states
                                    )
                    actions.append(agent.compute_action(agent_states))
                all_agents_states, rewards, done = env.step(actions)
    \end{lstlisting}
    
    A \textit{market} can be thought of as a stateful, repeated auction:
    \begin{itemize}
        \item A market is initialized with $m$ bidders, each of which has a state.
        \item A market lasts for $N$ days.
        \item Each day, auction items are put on sale. Each day, the bidders participate in an auction for the available items.
        \item Each bidder's state is updated at the end of every day. The state can track information such as auction items bought and amount spent.
    \end{itemize}

\bibliographystyle{plain}
\bibliography{references}

\end{document}

%% file: experiment_outcomes.tex
\begin{landscape}
\begingroup
\footnotesize
\setlength{\LTcapwidth}{\linewidth}
\setlength{\tabcolsep}{2pt}
\renewcommand{\arraystretch}{1.12}
\begin{longtable}{>{\raggedright\arraybackslash}p{2.80cm}>{\raggedright\arraybackslash}p{2.40cm}>{\raggedright\arraybackslash}p{2.40cm}>{\raggedright\arraybackslash}p{2.60cm}>{\raggedright\arraybackslash}p{1.90cm}>{\raggedright\arraybackslash}p{1.70cm}>{\raggedright\arraybackslash}p{2.60cm}>{\centering\arraybackslash}p{.60cm}>{\centering\arraybackslash}p{1.00cm}>{\centering\arraybackslash}p{.85cm}>{\centering\arraybackslash}p{.85cm}>{\centering\arraybackslash}p{.85cm}}
\caption{Qualitative outcomes by experiment configuration and seed. Each row specifies the algorithm, baseline or critic, initialization, hidden/output activations, optimizer, update setting, and training horizon. Codes are SH, sustained high; DC, delayed collapse; EC, early collapse; ZI, zero from initialization; FI, failed to reach informative reward; LL, limited/low reward; and NF, numerical failure. Operational rules are given in Appendix~\ref{sec:outcome_classification}; SH denotes a final-100-epoch mean reward of at least 5. The matched SGD--Adam comparison uses a common 1,000-epoch horizon; separate high-rate cohorts retain their listed horizons. Nominal learning rates are optimizer-specific settings.}\label{tab:rl_experiment_outcomes}\\
Experiment & Algorithm & Baseline /\newline critic & Initialization & Activation & Optimizer & Setting & $n$ & Epochs & Seed 123 & Seed 456 & Seed 789 \\
\hline
\endfirsthead
Experiment & Algorithm & Baseline /\newline critic & Initialization & Activation & Optimizer & Setting & $n$ & Epochs & Seed 123 & Seed 456 & Seed 789 \\
\hline
\endhead
Untreated initialization & REINFORCE & off & Glorot, uncentered & ReLU / ReLU & SGD & LR=1e-07 & 3 & 500 & ZI & ZI & ZI \\
Untreated initialization & REINFORCE & off & Glorot, uncentered & ReLU / ReLU & SGD & LR=1e-06 & 3 & 500 & ZI & ZI & ZI \\
Untreated initialization & REINFORCE & off & Glorot, uncentered & ReLU / ReLU & SGD & LR=1e-05 & 3 & 500 & ZI & ZI & ZI \\
Untreated initialization & REINFORCE & on & Glorot, uncentered & ReLU / ReLU & SGD & LR=1e-07 & 3 & 500 & ZI & ZI & ZI \\
Untreated initialization & REINFORCE & on & Glorot, uncentered & ReLU / ReLU & SGD & LR=1e-06 & 3 & 500 & ZI & ZI & ZI \\
Untreated initialization & REINFORCE & on & Glorot, uncentered & ReLU / ReLU & SGD & LR=1e-05 & 3 & 500 & EC & ZI & ZI \\
Untreated initialization & ActorCritic & state-value critic & Glorot, uncentered & ReLU / ReLU & SGD & LR=1e-09 & 3 & 500 & NF & ZI & ZI \\
Untreated initialization & ActorCritic & state-value critic & Glorot, uncentered & ReLU / ReLU & SGD & LR=1e-08 & 3 & 500 & NF & ZI & ZI \\
Untreated initialization & ActorCritic & state-value critic & Glorot, uncentered & ReLU / ReLU & SGD & LR=1e-07 & 3 & 500 & NF & ZI & ZI \\
Bias-centered Glorot & REINFORCE & off & Glorot with centered bias & ReLU / ReLU & SGD & LR=1e-07 & 3 & 500 & EC & DC & ZI \\
Bias-centered Glorot & REINFORCE & off & Glorot with centered bias & ReLU / ReLU & SGD & LR=1e-06 & 3 & 500 & FI & EC & ZI \\
Bias-centered Glorot & REINFORCE & off & Glorot with centered bias & ReLU / ReLU & SGD & LR=1e-05 & 3 & 500 & NF & NF & ZI \\
Bias-centered Glorot & REINFORCE & on & Glorot with centered bias & ReLU / ReLU & SGD & LR=1e-07 & 3 & 500 & EC & DC & ZI \\
Bias-centered Glorot & REINFORCE & on & Glorot with centered bias & ReLU / ReLU & SGD & LR=1e-06 & 3 & 500 & FI & EC & ZI \\
Bias-centered Glorot & REINFORCE & on & Glorot with centered bias & ReLU / ReLU & SGD & LR=1e-05 & 3 & 500 & NF & NF & ZI \\
Bias-centered Glorot & ActorCritic & state-value critic & Glorot with centered bias & ReLU / ReLU & SGD & LR=1e-09 & 3 & 500 & FI & SH & ZI \\
Bias-centered Glorot & ActorCritic & state-value critic & Glorot with centered bias & ReLU / ReLU & SGD & LR=1e-08 & 3 & 500 & FI & DC & ZI \\
Bias-centered Glorot & ActorCritic & state-value critic & Glorot with centered bias & ReLU / ReLU & SGD & LR=1e-07 & 3 & 500 & NF & EC & ZI \\
Control 2 & REINFORCE & off & Centered zero head & ReLU / ReLU & SGD & LR=1e-07 & 3 & 500 & EC & EC & EC \\
Control 2 & REINFORCE & off & Centered zero head & ReLU / ReLU & SGD & LR=1e-06 & 3 & 500 & EC & EC & LL \\
Control 2 & REINFORCE & off & Centered zero head & ReLU / ReLU & SGD & LR=1e-05 & 3 & 500 & FI & FI & FI \\
Control 2 & REINFORCE & on & Centered zero head & ReLU / ReLU & SGD & LR=1e-07 & 3 & 500 & EC & EC & EC \\
Control 2 & REINFORCE & on & Centered zero head & ReLU / ReLU & SGD & LR=1e-06 & 3 & 500 & EC & EC & LL \\
Control 2 & REINFORCE & on & Centered zero head & ReLU / ReLU & SGD & LR=1e-05 & 3 & 500 & FI & FI & FI \\
Control 2 & ActorCritic & state-value critic & Centered zero head & ReLU / ReLU & SGD & LR=1e-09 & 3 & 500 & SH & SH & SH \\
Control 2 & ActorCritic & state-value critic & Centered zero head & ReLU / ReLU & SGD & LR=1e-08 & 3 & 500 & DC & DC & DC \\
Control 2 & ActorCritic & state-value critic & Centered zero head & ReLU / ReLU & SGD & LR=1e-07 & 3 & 500 & EC & EC & EC \\
Smooth parameterization & REINFORCE & off & Centered zero head & ReLU / ReLU & SGD & LR=1e-07 & 3 & 500 & EC & EC & EC \\
Smooth parameterization & REINFORCE & off & Centered zero head & ReLU / ReLU & SGD & LR=1e-06 & 3 & 500 & EC & EC & LL \\
Smooth parameterization & REINFORCE & off & Centered zero head & ReLU / ReLU & SGD & LR=1e-05 & 3 & 500 & FI & FI & FI \\
Smooth parameterization & REINFORCE & on & Centered zero head & ReLU / ReLU & SGD & LR=1e-07 & 3 & 500 & EC & EC & EC \\
Smooth parameterization & REINFORCE & on & Centered zero head & ReLU / ReLU & SGD & LR=1e-06 & 3 & 500 & EC & EC & LL \\
Smooth parameterization & REINFORCE & on & Centered zero head & ReLU / ReLU & SGD & LR=1e-05 & 3 & 500 & FI & FI & FI \\
Smooth parameterization & ActorCritic & state-value critic & Centered zero head & ReLU / ReLU & SGD & LR=1e-09 & 3 & 500 & SH & SH & SH \\
Smooth parameterization & ActorCritic & state-value critic & Centered zero head & ReLU / ReLU & SGD & LR=1e-08 & 3 & 500 & DC & DC & DC \\
Smooth parameterization & ActorCritic & state-value critic & Centered zero head & ReLU / ReLU & SGD & LR=1e-07 & 3 & 500 & EC & EC & EC \\
Smooth parameterization & REINFORCE & off & Centered zero head & ReLU / softplus & SGD & LR=1e-07 & 3 & 500 & LL & SH & SH \\
Smooth parameterization & REINFORCE & off & Centered zero head & ReLU / softplus & SGD & LR=1e-06 & 3 & 500 & SH & SH & SH \\
Smooth parameterization & REINFORCE & off & Centered zero head & ReLU / softplus & SGD & LR=1e-05 & 3 & 500 & SH & SH & SH \\
Smooth parameterization & REINFORCE & on & Centered zero head & ReLU / softplus & SGD & LR=1e-07 & 3 & 500 & LL & LL & SH \\
Smooth parameterization & REINFORCE & on & Centered zero head & ReLU / softplus & SGD & LR=1e-06 & 3 & 500 & SH & SH & SH \\
Smooth parameterization & REINFORCE & on & Centered zero head & ReLU / softplus & SGD & LR=1e-05 & 3 & 500 & SH & SH & SH \\
Smooth parameterization & ActorCritic & state-value critic & Centered zero head & ReLU / softplus & SGD & LR=1e-09 & 3 & 500 & FI & FI & FI \\
Smooth parameterization & ActorCritic & state-value critic & Centered zero head & ReLU / softplus & SGD & LR=1e-08 & 3 & 500 & LL & FI & FI \\
Smooth parameterization & ActorCritic & state-value critic & Centered zero head & ReLU / softplus & SGD & LR=1e-07 & 3 & 500 & SH & SH & SH \\
Smooth parameterization & REINFORCE & off & Centered zero head & ELU / softplus & SGD & LR=1e-07 & 3 & 500 & SH & SH & SH \\
Smooth parameterization & REINFORCE & off & Centered zero head & ELU / softplus & SGD & LR=1e-06 & 3 & 500 & SH & SH & SH \\
Smooth parameterization & REINFORCE & off & Centered zero head & ELU / softplus & SGD & LR=1e-05 & 3 & 500 & SH & SH & SH \\
Smooth parameterization & REINFORCE & on & Centered zero head & ELU / softplus & SGD & LR=1e-07 & 3 & 500 & SH & LL & SH \\
Smooth parameterization & REINFORCE & on & Centered zero head & ELU / softplus & SGD & LR=1e-06 & 3 & 500 & SH & SH & SH \\
Smooth parameterization & REINFORCE & on & Centered zero head & ELU / softplus & SGD & LR=1e-05 & 3 & 500 & SH & SH & SH \\
Smooth parameterization & ActorCritic & state-value critic & Centered zero head & ELU / softplus & SGD & LR=1e-09 & 3 & 500 & FI & FI & FI \\
Smooth parameterization & ActorCritic & state-value critic & Centered zero head & ELU / softplus & SGD & LR=1e-08 & 3 & 500 & LL & FI & LL \\
Smooth parameterization & ActorCritic & state-value critic & Centered zero head & ELU / softplus & SGD & LR=1e-07 & 3 & 500 & SH & SH & SH \\
Fixed high-rate SGD & REINFORCE & off & Centered zero head & ReLU / softplus & SGD & LR=3e-04 & 3 & 200 & FI & FI & FI \\
Fixed high-rate SGD & REINFORCE & off & Centered zero head & ReLU / softplus & SGD & LR=1e-03 & 3 & 200 & FI & FI & NF \\
Fixed high-rate SGD & REINFORCE & on & Centered zero head & ReLU / softplus & SGD & LR=3e-04 & 3 & 200 & NF & FI & FI \\
Fixed high-rate SGD & REINFORCE & on & Centered zero head & ReLU / softplus & SGD & LR=1e-03 & 3 & 200 & NF & FI & NF \\
Fixed high-rate SGD & ActorCritic & state-value critic & Centered zero head & ReLU / softplus & SGD & LR=3e-04 & 3 & 200 & FI & NF & FI \\
Fixed high-rate SGD & ActorCritic & state-value critic & Centered zero head & ReLU / softplus & SGD & LR=1e-03 & 3 & 200 & FI & EC & FI \\
Fixed high-rate SGD & ActorCritic & state-value critic & Centered zero head & ReLU / softplus & SGD & LR=3e-03 & 3 & 200 & FI & FI & FI \\
Fixed high-rate Adam & REINFORCE & off & Centered zero head & ReLU / softplus & Adam & LR=3e-04 & 3 & 1000 & SH & LL & SH \\
Fixed high-rate Adam & REINFORCE & off & Centered zero head & ReLU / softplus & Adam & LR=1e-03 & 3 & 1000 & SH & SH & SH \\
Fixed high-rate Adam & REINFORCE & on & Centered zero head & ReLU / softplus & Adam & LR=3e-04 & 3 & 1000 & SH & LL & SH \\
Fixed high-rate Adam & REINFORCE & on & Centered zero head & ReLU / softplus & Adam & LR=1e-03 & 3 & 1000 & SH & SH & SH \\
Fixed high-rate Adam & ActorCritic & state-value critic & Centered zero head & ReLU / softplus & Adam & LR=3e-04 & 3 & 1000 & DC & SH & SH \\
Fixed high-rate Adam & ActorCritic & state-value critic & Centered zero head & ReLU / softplus & Adam & LR=1e-03 & 3 & 1000 & DC & SH & DC \\
Fixed high-rate Adam & ActorCritic & state-value critic & Centered zero head & ReLU / softplus & Adam & LR=3e-03 & 3 & 1000 & EC & DC & EC \\
Matched fixed-rate SGD vs. Adam & ActorCritic & state-value critic & Centered zero head & ReLU / softplus & Adam & LR=3e-04 & 3 & 1000 & DC & SH & SH \\
Matched fixed-rate SGD vs. Adam & ActorCritic & state-value critic & Centered zero head & ReLU / softplus & Adam & LR=1e-03 & 3 & 1000 & DC & SH & DC \\
Matched fixed-rate SGD vs. Adam & ActorCritic & state-value critic & Centered zero head & ReLU / softplus & Adam & LR=3e-03 & 3 & 1000 & EC & DC & EC \\
Matched fixed-rate SGD vs. Adam & ActorCritic & state-value critic & Centered zero head & ReLU / softplus & SGD & LR=1e-07 & 3 & 1000 & SH & SH & SH \\
Matched fixed-rate SGD vs. Adam & ActorCritic & state-value critic & Centered zero head & ReLU / softplus & SGD & LR=1e-06 & 3 & 1000 & DC & DC & DC \\
Matched fixed-rate SGD vs. Adam & ActorCritic & state-value critic & Centered zero head & ReLU / softplus & SGD & LR=1e-05 & 3 & 1000 & EC & EC & EC \\
Matched fixed-rate SGD vs. Adam & REINFORCE & off & Centered zero head & ReLU / softplus & Adam & LR=3e-04 & 3 & 1000 & SH & LL & SH \\
Matched fixed-rate SGD vs. Adam & REINFORCE & off & Centered zero head & ReLU / softplus & Adam & LR=3e-03 & 3 & 1000 & SH & SH & SH \\
Matched fixed-rate SGD vs. Adam & REINFORCE & off & Centered zero head & ReLU / softplus & Adam & LR=1e-02 & 3 & 1000 & SH & SH & EC \\
Matched fixed-rate SGD vs. Adam & REINFORCE & off & Centered zero head & ReLU / softplus & SGD & LR=1e-07 & 3 & 1000 & SH & SH & SH \\
Matched fixed-rate SGD vs. Adam & REINFORCE & off & Centered zero head & ReLU / softplus & SGD & LR=1e-05 & 3 & 1000 & SH & SH & SH \\
Matched fixed-rate SGD vs. Adam & REINFORCE & off & Centered zero head & ReLU / softplus & SGD & LR=1e-04 & 3 & 1000 & EC & DC & FI \\
Matched fixed-rate SGD vs. Adam & REINFORCE & on & Centered zero head & ReLU / softplus & Adam & LR=3e-04 & 3 & 1000 & SH & LL & SH \\
Matched fixed-rate SGD vs. Adam & REINFORCE & on & Centered zero head & ReLU / softplus & Adam & LR=3e-03 & 3 & 1000 & SH & SH & SH \\
Matched fixed-rate SGD vs. Adam & REINFORCE & on & Centered zero head & ReLU / softplus & Adam & LR=1e-02 & 3 & 1000 & SH & SH & LL \\
Matched fixed-rate SGD vs. Adam & REINFORCE & on & Centered zero head & ReLU / softplus & SGD & LR=1e-07 & 3 & 1000 & SH & SH & SH \\
Matched fixed-rate SGD vs. Adam & REINFORCE & on & Centered zero head & ReLU / softplus & SGD & LR=1e-05 & 3 & 1000 & SH & SH & SH \\
Matched fixed-rate SGD vs. Adam & REINFORCE & on & Centered zero head & ReLU / softplus & SGD & LR=1e-04 & 3 & 1000 & FI & NF & FI \\
Isolated adaptive step size & REINFORCE & off & Centered zero head & ReLU / ReLU & adaptive SGD & epsilon=1e-08 & 3 & 1000 & SH & SH & SH \\
Isolated adaptive step size & REINFORCE & off & Centered zero head & ReLU / ReLU & adaptive SGD & epsilon=1e-07 & 3 & 1000 & SH & SH & SH \\
Isolated adaptive step size & REINFORCE & off & Centered zero head & ReLU / ReLU & adaptive SGD & epsilon=1e-06 & 3 & 1000 & SH & SH & SH \\
Isolated adaptive step size & REINFORCE & on & Centered zero head & ReLU / ReLU & adaptive SGD & epsilon=1e-08 & 3 & 1000 & SH & SH & SH \\
Isolated adaptive step size & REINFORCE & on & Centered zero head & ReLU / ReLU & adaptive SGD & epsilon=1e-07 & 3 & 1000 & SH & SH & SH \\
Isolated adaptive step size & REINFORCE & on & Centered zero head & ReLU / ReLU & adaptive SGD & epsilon=1e-06 & 3 & 1000 & SH & SH & SH \\
Isolated adaptive step size & ActorCritic & state-value critic & Centered zero head & ReLU / ReLU & adaptive SGD & epsilon=1e-08 & 3 & 1000 & LL & SH & SH \\
Isolated adaptive step size & ActorCritic & state-value critic & Centered zero head & ReLU / ReLU & adaptive SGD & epsilon=1e-07 & 3 & 1000 & SH & SH & SH \\
Isolated adaptive step size & ActorCritic & state-value critic & Centered zero head & ReLU / ReLU & adaptive SGD & epsilon=1e-06 & 3 & 1000 & DC & DC & SH \\
Combined treatment & REINFORCE & off & Centered zero head & ReLU / softplus & adaptive SGD & epsilon=3e-06 & 3 & 1000 & SH & SH & SH \\
Combined treatment & REINFORCE & off & Centered zero head & ReLU / softplus & adaptive SGD & epsilon=1e-05 & 3 & 1000 & SH & SH & SH \\
Combined treatment & REINFORCE & off & Centered zero head & ReLU / softplus & adaptive SGD & epsilon=3e-05 & 3 & 1000 & SH & SH & SH \\
Combined treatment & REINFORCE & on & Centered zero head & ReLU / softplus & adaptive SGD & epsilon=3e-06 & 3 & 1000 & SH & LL & SH \\
Combined treatment & REINFORCE & on & Centered zero head & ReLU / softplus & adaptive SGD & epsilon=1e-05 & 3 & 1000 & SH & SH & SH \\
Combined treatment & REINFORCE & on & Centered zero head & ReLU / softplus & adaptive SGD & epsilon=3e-05 & 3 & 1000 & SH & SH & SH \\
Combined treatment & ActorCritic & state-value critic & Centered zero head & ReLU / softplus & adaptive SGD & epsilon=3e-06 & 3 & 1000 & SH & SH & SH \\
Combined treatment & ActorCritic & state-value critic & Centered zero head & ReLU / softplus & adaptive SGD & epsilon=1e-05 & 3 & 1000 & DC & SH & DC \\
Combined treatment & ActorCritic & state-value critic & Centered zero head & ReLU / softplus & adaptive SGD & epsilon=3e-05 & 3 & 1000 & DC & DC & DC \\
\end{longtable}
\endgroup
\end{landscape}